\documentclass[journal=jacsat,manuscript=article]{achemso}

\usepackage[version=3]{mhchem} 
\usepackage{amsmath,amssymb,amsfonts}
\usepackage{algorithmic}
\usepackage{graphicx}
\usepackage{textcomp}
\usepackage{xcolor}
\usepackage{subcaption}
\usepackage[font=small,labelfont=bf]{caption}
\usepackage{hyperref}

\author{Akshay Raman}
\altaffiliation{These authors contributed equally.}
\email{akshayra@andrew.cmu.edu}

\author{Chad Merrill}
\altaffiliation{These authors contributed equally.}
\email{camerril@alumni.cmu.edu}

\author{Abraham George}
\email{aigeorge@andrew.cmu.edu}

\author{Amir Barati Farimani}
\email{barati@cmu.edu}

\affiliation{
  Department of Mechanical Engineering, Carnegie Mellon University, 5000 Forbes Ave, Pittsburgh,
  PA 15217, USA
}

\title[An \textsf{achemso} demo]
  {LLM-Drone: Aerial Additive Manufacturing with Drones Planned Using Large Language Models}

\begin{document}


\begin{abstract}
Additive manufacturing (AM) has transformed the production landscape by enabling the precision creation of complex geometries. However, AM faces limitations when applied to challenging environments, such as elevated surfaces and remote locations. Aerial additive manufacturing, facilitated by drones, presents a solution to these challenges by allowing construction in previously inaccessible areas. However, despite advances in methods for the planning, control, and localization of drones, the accuracy of these methods is insufficient to run traditional feedforward extrusion-based additive manufacturing processes (such as Fused Deposition Manufacturing). Recently, the emergence of LLMs has revolutionized various fields by introducing advanced semantic reasoning and real-time planning capabilities. This paper proposes the integration of LLMs with aerial additive manufacturing to assist with the planning and execution of construction tasks, granting greater flexibility and enabling a feed-back based design and construction system. Using the semantic understanding and adaptability of LLMs, we can overcome the limitations of drone based systems by dynamically generating and adapting building plans on site, ensuring efficient and accurate construction even in constrained environments. We propose a novel methodology that leverages LLMs to design a construction plan for drone-based manufacturing, and adjust the plan in real time to address any errors that may occur. Our system is able to design and build structures given only a semantic prompt and has shown success in understanding the spatial environment despite tight planning constraints. Our method's feedback system enables replanning using the LLM if the manufacturing process encounters unforeseen errors, without requiring complicated heuristics or evaluation functions. Combining the semantic planning with automatic error correction, our system achieved a 90\% build accuracy, converting simple text prompts to build structures. 
\end{abstract}

\section{Introduction}
Additive manufacturing (AM) has revolutionized production by enabling the fabrication of complex geometries and custom components with high precision and minimal material waste \cite{jadhav2024generative, jadhav2024large, jadhav2024llm}. Its applications span diverse industries, from aerospace to healthcare, where design flexibility and efficient material usage are critical \cite{uriondo2015present, ahangar2019current}. However, conventional AM technologies are inherently limited by the need for stable, controlled environments and fixed build platforms. This reliance on stationary printing systems restricts their deployment in settings where structures need to be assembled in place, or where the build area is a dynamic or unstructured environment. 

To address these spatial limitations, drones have been proposed as a solution to extend the reach of additive manufacturing \cite{6853477, Latteur2015DroneBasedAM}. Their mobility and navigational adaptability across diverse terrains present an opportunity to overcome location-based constraints. However, recent approaches that integrate drones with traditional additive manufacturing methods have encountered significant hurdles. The inherent instability of drones in flight poses difficulties in achieving the precision required for layer-by-layer fabrication, particularly in fluid-to-solid transitions commonly used in additive manufacturing processes \cite{zhang2022aerial}.

Recognizing these limitations, we introduce a novel approach that reimagines drone-assisted manufacturing. Rather than relying on drones to perform traditional additive manufacturing tasks, our method uses predefined geometries, specifically magnetically connecting blocks, that drones can transport and assemble. Using these standardized building blocks, we eliminate the need for precise in-flight material deposition, alleviating the challenges of drone stability and control.

Beyond addressing these physical constraints, we leverage Large Language Models (LLMs) to enhance the design and planning phases of the process. Much like a slicer in traditional 3D printing, the LLM translates high-level design goals into structured, executable plans for our drone-based manufacturing system. With their advanced reasoning and creative capabilities, LLMs can dynamically generate and adapt designs based on the current build state and user-defined requests. By combining predefined geometries with intelligent design planning, our approach not only enhances construction accuracy but also enables a flexible and scalable framework for drone-assisted manufacturing across diverse environments.

To implement this approach, we propose the LLM-Drone pipeline, which consists of three key modules that work together to enable drone-assisted manufacturing. The LLM Planning Module generates structured designs based on the current 3D build state and user-defined requests, much like a slicer in traditional 3D printing. The Computer Vision Module aligns coordinate systems and estimates the build location’s current state to ensure accurate block placement. Finally, the Mechanical Module consists of the drone and magnetically connecting blocks, allowing for precise pick-and-place assembly. By integrating predefined geometries with intelligent design planning, this pipeline ensures accuracy in construction while enabling a flexible and scalable framework for drone-assisted manufacturing across diverse environments.

\begin{figure}[!]
    \centering
    \includegraphics[width=\textwidth]{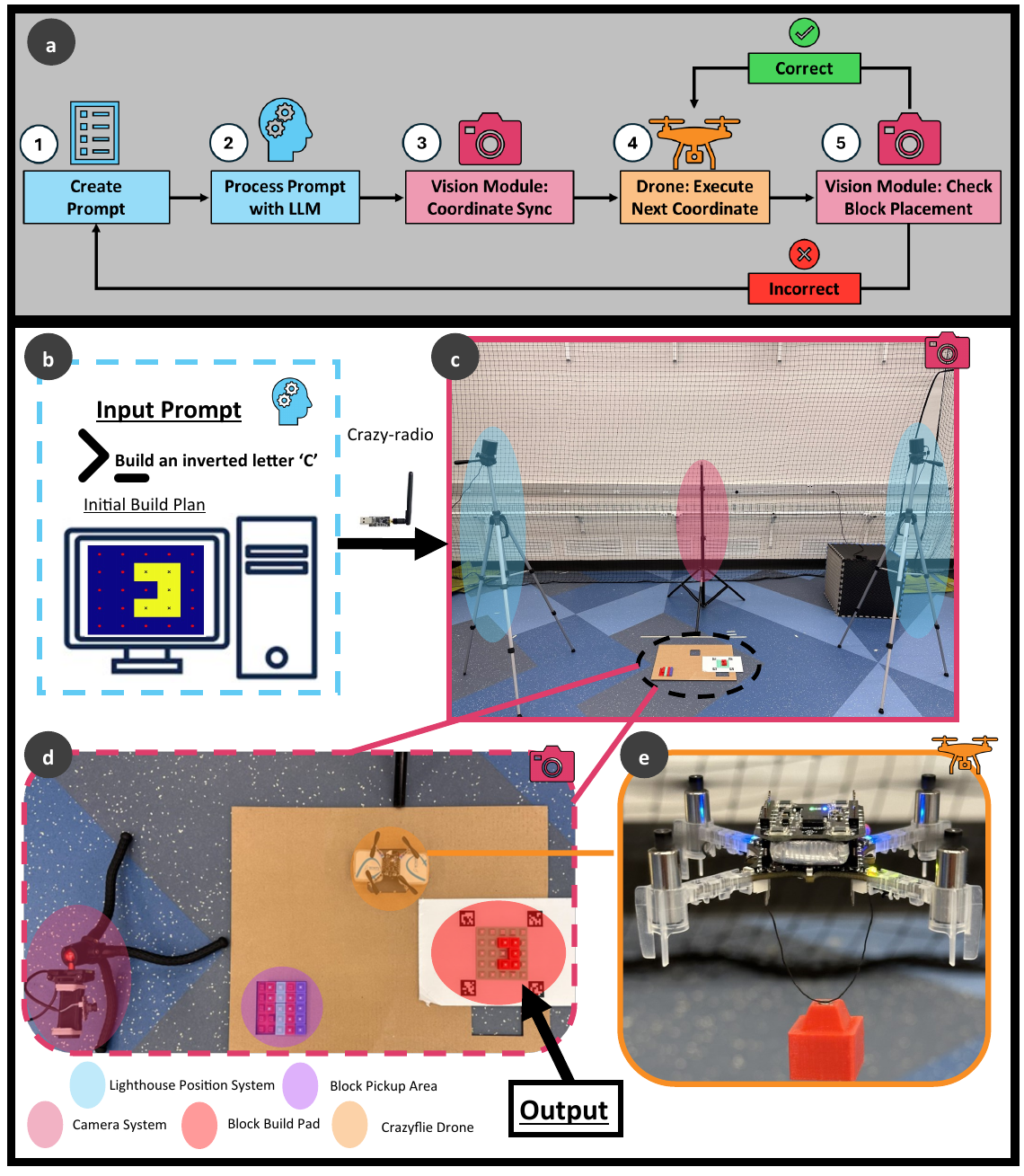}
    \caption{System Overview. (a) The main modules required for the additive manufacturing process. (b) A prompt is created from a default prompt template that includes current scene info and a design request. The LLM processes the prompt and outputs the coordinates needed to achieve the design. (c) The vision module aligns the Crazyflie coordinates with the LLM output coordinates using the Crazyflie lighthouse. (d) The drone places a block and the vision model verifies the placement. If incorrect, the current scene is passed back to (b) for remprompt and a new set of coordinates is generated to finish the design. (e) The Crazyflie drone transporting a building block from pickup to drop off.}
    \label{fig:A_System_Overview}
\end{figure}

\section{Related Work}
\subsection{Aerial Additive Manufacturing}
Aerial additive manufacturing (AAM) has gained significant attention in recent years as researchers explore novel ways to integrate autonomous aerial vehicles, such as drones, into the construction and manufacturing processes. The primary challenge in AAM lies in achieving precise material deposition and structural assembly in dynamic and often unstructured environments.

Early efforts in drone-based construction focused on cooperative assembly, where multiple autonomous drones collaborated to construct architectural structures. Augugliaro et al. pioneered the concept of cooperative aerial construction using quadrotors, demonstrating the ability of drones to assemble structures with predefined geometric constraints \cite{6853477}. Later, Latteur et al. extended this work by investigating the feasibility of drone-based additive manufacturing for architectural applications, emphasizing the need for robust localization and control mechanisms \cite{Latteur2015DroneBasedAM}.

Recent advancements have sought to improve the accuracy of drone-based additive manufacturing by integrating perception and control systems. Zhang et al. introduced a multi-drone approach to aerial 3D printing, where autonomous robots collaboratively extruded materials to construct structures \cite{zhang2022aerial}. Their work demonstrated the potential of using real-time feedback loops to correct errors in flight. However, challenges such as vibration-induced inaccuracies and the need for precise material deposition remain significant barriers.

\subsection {Integration of LLMs into Robotic Planning}
The surge of Large Language Models (LLMs) has brought forth new possibilities for more flexible and semantic-based planning in robotics. Early works, such as SayCan \cite{ahn2022can}, an LLM-based framework for instructing household robots in real time using a combination of language understanding and probabilistic checks to ensure feasible action sequences, ushered in a golden age for robotic advancements using LLMs. Following works have shown that LLMs can be used in a variety of robotic applications, ranging from tool use in long horizon tasks \cite{car2024plato}, to deformable object manipulation \cite{bartsch2024llm}, to additive-manufacturing specific applications such as automated parameter optimization for 3D printers \cite{jadhav2024llm}. 

A key benefit of coupling LLMs with robot platforms, which our work seeks to leverage, is the ability to perform context-aware error correction. For instance, Huang et al. demonstrated that an LLM-based planning system can revise multi-step manipulation procedures after encountering unexpected disturbances, identifying partial successes or anomalies in the environment and proposing new step sequences to ensure task completion \cite{huang2022inner}. Vemprala et al. similarly highlighted how LLMs can serve as an effective high-level planner for mobile and tabletop robotic tasks, dynamically updating instructions based on real-time sensor feedback \cite{vemprala2024chatgpt}. Liang et al. extended these ideas by showing that “Code as Policies” enables language models to both interpret human commands and generate executable robot actions, including adaptations mid-task if the environment deviates from initial assumptions \cite{liang2023code}. While most prior work in LLM-based planning has centered on ground or table-top robotics, the same logic applies to aerial platforms: the language model can ingest sensor or vision data about a building site, and re-plan if a drone fails to place a block accurately.

\subsection{Bitcraze Crazyflie Platform}
This work explores additive manufacturing using the Bitcraze Crazyflie platform \cite{bitcraze_documentation}, specifically the Crazyflie 2.1 nanoquadcopter - a small and inexpensive quadcopter designed for research and prototyping. For our research, we specifically use the Motion Commander module in the provided Python API, which allows for precise control of the drone’s position and movement. Communication with the drone is handled via the CrazyRadio module, while the integrated Lighthouse positioning system ensures accurate tracking in three-dimensional space. This combination of Python-based API, reliable communication, precise positioning, and low cost makes the Crazyflie 2.1 an ideal choice for academic research. This applicability is evidenced by the hardware's prior research applications, ranging from swarm coordination \cite{preiss2017crazyswarm}, to autonomous navigation and mapping \cite{mcguire2019minimal}, to multi-agent deep reinforcement learning \cite{tadevosyan2025attentionswarm}, to spacecraft formation simulation \cite{de2024testing}.

\section{Environment Setup}
Our experimental setup up consists of a 3D environment represented within the world as $W \subseteq \mathbb{R}^3$ with size H*W*L, discretized into 1 cm sections. This array delineates the environmental layout, classifying regions as obstacles or free space, which facilitates the navigation of drones. A roadmap can be constructed within this environment to traverse free space while avoiding obstacles during the construction process.

The experimental infrastructure utilizes Bitcraze's Crazyflie ecosystem, including the Crazyflie Client, Crazyradio, and Lighthouse System, as illustrated in Figures \ref{fig:A_System_Overview}b and \ref{fig:A_System_Overview}c. This setup was used consistently across all data collection phases. Figures \ref{fig:A_System_Overview}c, \ref{fig:A_System_Overview}d and \ref{fig:A_System_Overview}e highlight additional hardware including a Crazyflie 2.1 drone, a 1080p webcam, and the 3D printed blocks and building pads designed for the experiments. The Crazyflie ecosystem was selected for its robust built-in API, which seamlessly integrates with both the controller and the autonomous flight controller. This integration allowed our team to concentrate on developing a robust additive manufacturing system.

\section{Methodology}

The LLM-Drone pipeline \cite{chen2023scalable} integrates our LLM Planning Module, Computer Vision Module, and Mechanical Module to autonomously construct a user-defined design following an initial request. Figure \ref{fig:A_System_Overview}a illustrates how these modules work together to enable a drone-based additive manufacturing process. First, the user’s design request is captured and passed into a default prompt that includes additional instructions. This prompt is then input into an LLM, which generates a JSON script containing a set of coordinates as an action array for completing the build task. Next, the Computer Vision module converts these coordinates into the drone’s world coordinate system using the coordinate synchronization algorithm. This algorithm converts the integer coordinates output by the LLM to the real-world positions of those locations in the scene. The converted coordinate array is then sent to the Crazyflie client, and the onboard controller executes the block pickup and dropoff commands. Afterward, the Computer Vision module verifies the block’s placement. If the block is in the correct location, the drone proceeds with the next build action. If the location is incorrect, a new prompt is generated using observations of the environment, including the current position of any placed blocks. The LLM processes this updated scene and generates new coordinates to continue the design from the current state. Further details on each of the three major modules are provided in the remainder of this section.

\subsection{LLM Planning Module}

\begin{figure}[tb!]
\centerline{\includegraphics[width=\textwidth]{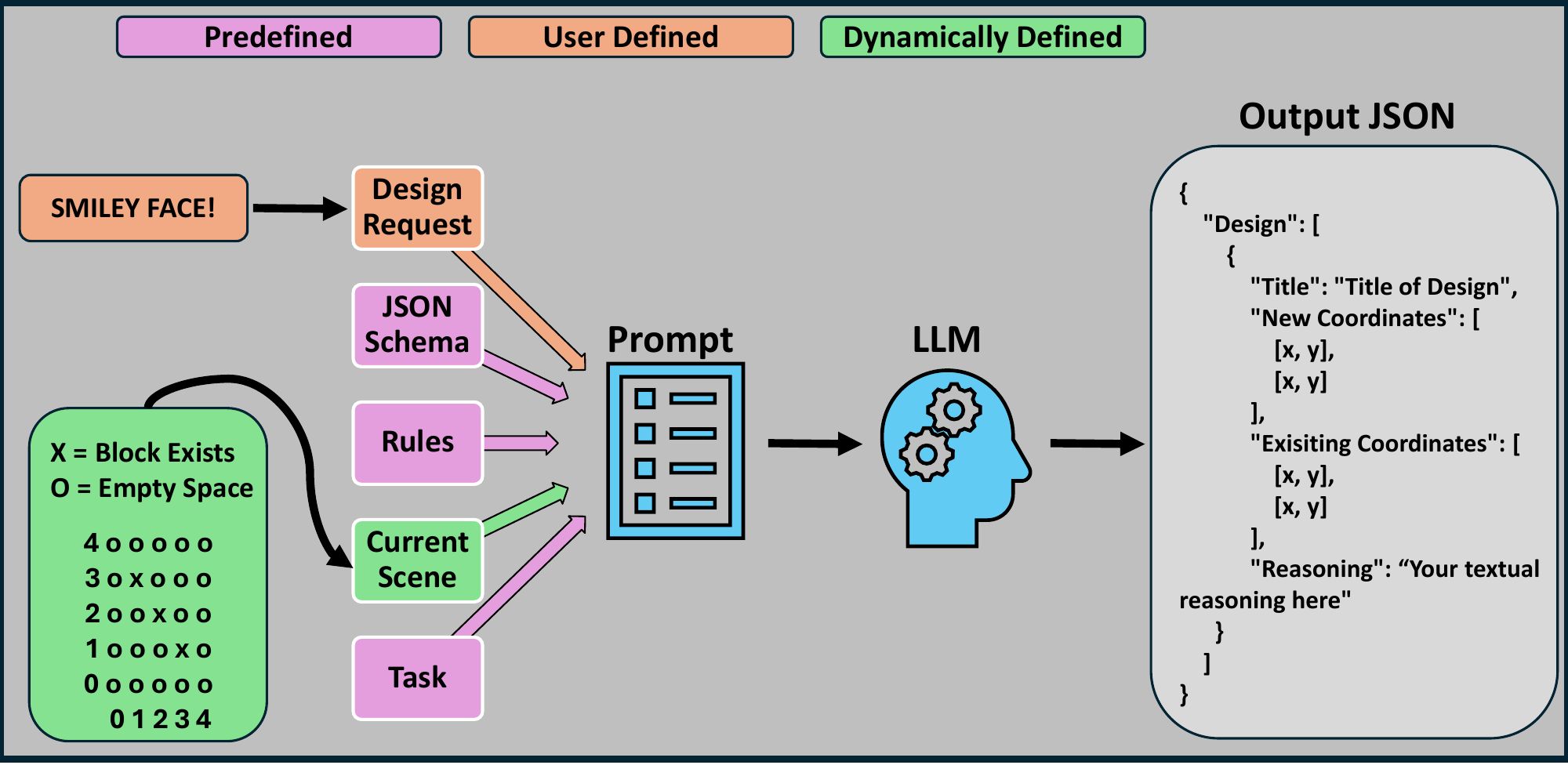}}
\caption{The prompt is broken into 5 parts: Design request, JSON Schema, Rules, Current Scene, and Task. The Task, Rules, and JSON Schema are predefined and do not change. The Design Request is input by the user at the start of the build process. The Current Scene is captured and presented each time a prompt is called.}
\label{fig:Prompt_Structure}
\end{figure}

The LLM Planning Module includes the creation of the prompt and the processing of that prompt by an LLM. Figure \ref{fig:Prompt_Structure} shows the structure for both the initial prompt and re-prompts, and Appendix A shows an example LLM prompt and response. The basic prompt structure remains the same for both types of prompts. The main components of the prompt are highlighted below.
\begin{itemize}
    \item \textbf{Task:} The requirements of the task and what needs to be done. In our case, this is the definition of the coordinates that a drone will execute to build a desired design.
    \item \textbf{Design Request:} A text input from the user describing their desired design. 
    \item \textbf{Current Scene:} This section begins by defining the dimensions and layout of the build space to the LLM. This includes a definition of the pad coordinate system. We then provide a textual representation of the grid system where ``o" represents free spaces and ``x" represents spaces that a block already occupies, which is generated each time a prompt is created. During the drone execution phase, if an error is detected in the dropoff position by the Vision Module, this textual representation of the build space is updated in the reprompt. This ensures that the LLM is prompted to redesign the required task using the existing block states as the initialization point.
    \item \textbf{Rules:} Defines rules the LLM must follow when selecting coordinates to create the requested design. 
    \item \textbf{Output Schema:} Defines the JSON output schema with which the LLM should respond. The structure includes a title for the design, a set of coordinates for the drone to execute, any coordinates currently utilized on the build plate, and the LLMs' reasoning on how it created the design. 
\end{itemize}
\subsection{Computer Vision Module}

The LLM-Drone Computer Vision Module integrates three submodules to enhance coordination and precision in drone operations. This section outlines the reasoning behind each vision state check and its integration within our application.

\subsubsection{Coordinate Synchronization}

\begin{figure}[h!t]
\centerline{\includegraphics[width=\textwidth]{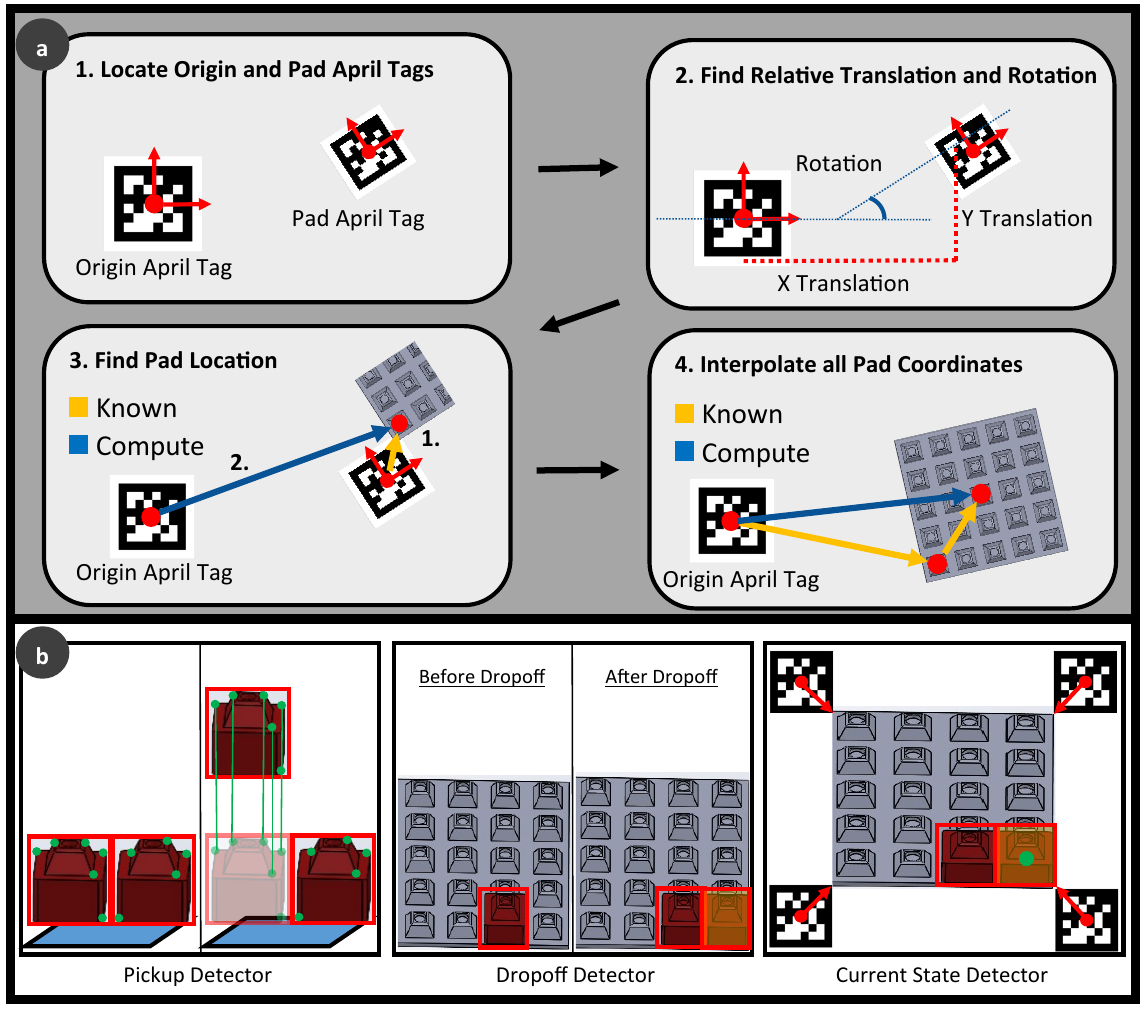}}
\caption{(a) Coordinate Sync algorithm overview. 1) Locate the origin AprilTag and the pad AprilTags. 2) Find the relative transformation between the origin and pad AprilTags. 3) Calculate the pose of the lower corner of the pad wrt. the origin. 4) Interpolate any pad coordinate by adding known vector from origin to bottom left position of pad. (b) The vision module's 3 main purposes. Pickup Detector: Green points represent corner points for Lucas-Kanade to follow. If the average of all tracking points within the YOLO bounding box is greater than the threshold (as shown on the right), the block is deemed as moved. Dropoff Detector: A frame is captured before the drone enters the region and drops a block. Background subtraction is used to compare with the initial image to determine if a block has been placed. Current State Detector: The corners of the pad are interpolated from the location of the AprilTags. The center of spatial subtraction is given by the green dot. The dot is interpolated to the closest pad location in x and y using AprilTags.}
\label{fig:C_Vision_Module}
\end{figure}

To minimize computational load on the LLM and prioritize overall shape construction, the LLM returns instructions in the build pad's local coordinate frame. The Coordinate Sync module translates these local coordinates from the LLM frame to the world coordinate system. In our setup, we use the Crazyflie's Lighthouse System to define the world coordinate frame \cite{evans2001rotations}. 

We propose two methods to determine the relationship between the pad's frame and the world coordinate frame:
\begin{enumerate}
\item Hardcoded Method: We fix the location of the bottom right point on the pad and interpolate all other coordinates based on rotation and local positioning. This method provides higher accuracy but requires that the build position be predetermined.

\item AprilTag-Based Method: We use AprilTags to compute the relationship between the world and the pad frames. This method offers flexibility in placing the pad at any location, allowing real-time computation of its position.
\end{enumerate}

Figure \ref{fig:C_Vision_Module} illustrates how coordinate synchronization aligns the local pad coordinates with the world frame and converts the LLM output action matrix into world coordinates.

\textbf{Locate Necessary AprilTags:} Figure \ref{fig:C_Vision_Module} shows the two AprilTags detected by the calibrated stationary camera. These tags are used to establish their positions in the camera's coordinate frame, $C_f$ \cite{kallwies2020determining}.

\textbf{Find Relative Translation and Rotation:} 
Since the detected tags are represented in the camera frame, $C_f$, we compute the relative translation and rotation of the pad's AprilTag with respect to the origin AprilTag. The AprilTag origin is aligned with the world frame. Thus, the relative translation vector ($t_ {rel}$) and the relative rotation matrix ($R_{rel}$) in the world frame, $W_f$, can be calculated as follows:
\begin{equation}
 t_{rel}=P_{origin_{C_f}}-P_{pad_{C_f}} \label{eq}
\end{equation}
\begin{equation}
R_{rel}=R_{W_fC_f}*R_{C_fP_f}=R_{C_fW_f}^{-1}*R_{C_fP_f}  \label{eq}
\end{equation}

\textbf{Compute Pad Location:} Using the position and rotation of the April tag, and the known position of the pad coordinate frame origin (the first notch) wrt. the April tag, the location of this notch can be calculated \cite{springer2022evaluation, kalaitzakis2021fiducial}. The location of this coordinate in the world frame is computed as:
\begin{equation}
P_{notch1_{W_f}}=t_{rel}+R_{rel}*P_{notch1_{P_f}} \label{eq}
\end{equation}

\textbf{Interpolate All Pad Points:} The location of each pad coordinate (represented by the LLM as a pair of indices from (0,0) (bottom left) to (environment size, environment size) (top right) can be interpolated based on the known location of the bottom left corner of the pad in the world coordinate frame. Figure \ref{fig:C_Vision_Module} shows the computation required to find the pad coordinate (2,2) in world coordinates. The computed coordinates are sent to the Crazyflie controller for execution.


\subsubsection{Verify Action Completion and State Correctness}
As depicted in Figure \ref{fig:C_Vision_Module}, the vision model includes verification of the completion of the pickup and drop-off action, as well as the current placement of the block within the build pad (correctness of state). Figure \ref{fig:C_Vision_Module} shows the architecture behind the verifier.

\textbf{Pickup Detector:} YOLO-v8 is trained with images of the construction blocks \cite{terven2023comprehensive, wang2023uav}. The blocks are 3D printed with various colors to illustrate that our vision system is robust to alterations in lighting conditions and block appearance. A camera observes the manufacturing process and a YOLO-v8 model is used to track all building blocks in the frame. An ID is associated with each tracked block and the bounding box for each tracked block is determined \cite{xianbao2021improved}. Within each bounding box, the features of the top corner of the block are detected and their movement is tracked using Lucas-Kanade, as shown in Figure \ref{fig:C_Vision_Module} \cite{baker2004lucas}. Let ${p}_i^{(t)}$ represent the position of the $i$-th feature point at time $t$. The vertical displacement of this point between frames $t$ and $t+3$ is given by: \[ \Delta y_i = p_{i,y}^{(t+3)} - p_{i,y}^{(t)} \] where $p_{i,y}$ is the $y$-coordinate of the $i$-th point. The total vertical displacement $\Delta Y$ is the sum of the vertical displacements of all points: \[ \Delta Y = \sum_{i=1}^{N} \Delta y_i = \sum_{i=1}^{N} \left( p_{i,y}^{(t+3)} - p_{i,y}^{(t)} \right) \] where $N$ is the number of feature points. Define a pickup threshold $T$. If $\Delta Y$ exceeds this threshold, determine that the block has been picked up: \[ \text{Pickup} = \begin{cases} \text{True} & \text{if } \Delta Y > T \\ \text{False} & \text{otherwise} \end{cases} \]

\textbf{Dropoff Detector:} An overview camera was used to observe the construction area, both to determine if a block had been deposited (Dropoff Detector) and to verify the state of the construction (Current State Detector). The bounds of the pad are estimated using the AprilTags on all four corners of the pad. YOLO-v8 is applied to the region of the image where the pad is located (as determined by the April Tags) to track the blocks \cite{terven2023comprehensive}. If after placing a block, the number of blocks that YOLO-v8 detects in the scene increases, then a block has been successfully dropped. However, sometimes YOLO-v8 is inconsistent. To verify that a block has been added, a frame before the drone appears and attempts to drop a block is captured, as well as a frame after the drone attempts a dropoff and clears the space. These two frames are subtracted, and the resulting difference frame is examined to verify that the changed pixels correspond with the recorded block addition. This method is applied after each block placement, as multiple concurrent additions may lead to a block being missed by YOLO\cite{xiong2024combining, garcia2020background}.

\textbf{Current State Detector:} To determine the position of a block on the pad in the pad coordinate system and also to know whether a block is stacked or placed behind another block, we perform the following steps: \subsection*{1. Pad Coordinate System Mapping} Detect the corners of the pad using AprilTags. The pad corners were detected and ordered as follows: \( \mathbf{p}_0 \) (top-left), \( \mathbf{p}_1 \) (top-right), \( \mathbf{p}_2 \) (bottom-left), and \( \mathbf{p}_3 \) (bottom-right). Using these corners, map the block's position to the pad coordinate system. \subsection*{2. Background Subtraction and Center of Change} Let \( I_{\text{before}} \) and \( I_{\text{after}} \) be the images before and after dropping the block. The difference image \( D \) is calculated as: \[ D = \left| I_{\text{after}} - I_{\text{before}} \right| \] Identify the region of maximum change in \( D \). Let \( \Omega \) be the set of all pixels in this region \cite{kumar2016video}. The centroid \( (c_x, c_y) \) of this region is given by: \[ c_x = \frac{1}{|\Omega|} \sum_{(x,y) \in \Omega} x , c_y = \frac{1}{|\Omega|} \sum_{(x,y) \in \Omega} y \] \subsection*{3. Interpolating Pad Grid Coordinates} Using bilinear interpolation, calculate the grid coordinate: \[ x_{\text{grid}} = \left\lfloor \frac{c_x - p_{0,x}}{p_{1,x} - p_{0,x}} \cdot pad\_size \right\rfloor \] \[ y_{\text{grid}} = \left\lfloor \frac{c_y - p_{0,y}}{p_{2,y} - p_{0,y}} \cdot pad\_size \right\rfloor \] \subsection*{4. Determine if Stacked or Placed Behind} Check if the new block is placed in a previously occupied position. Let \( \epsilon_d \) be a distance threshold for considering the positions to be the same. Compute: \[ d_i = \| \mathbf{p}_{\text{new}} - \mathbf{p}_i \| \] If \( \min(d_i) < \epsilon_d \), check if the region of change is small. Let \( A \) be the area of the region of change and \( \epsilon_A \) be the area threshold: \[ \text{Stacked} \text{ if } \min(d_i) < \epsilon_d \text{ and } A < \epsilon_A \] \[ \text{Placed Behind} \text{ if } \min(d_i) > \epsilon_d \text{ or } A \geq \epsilon_A \]
\subsection{Block Design}

\begin{figure}[h!t]
\centerline{\includegraphics[width=\textwidth]{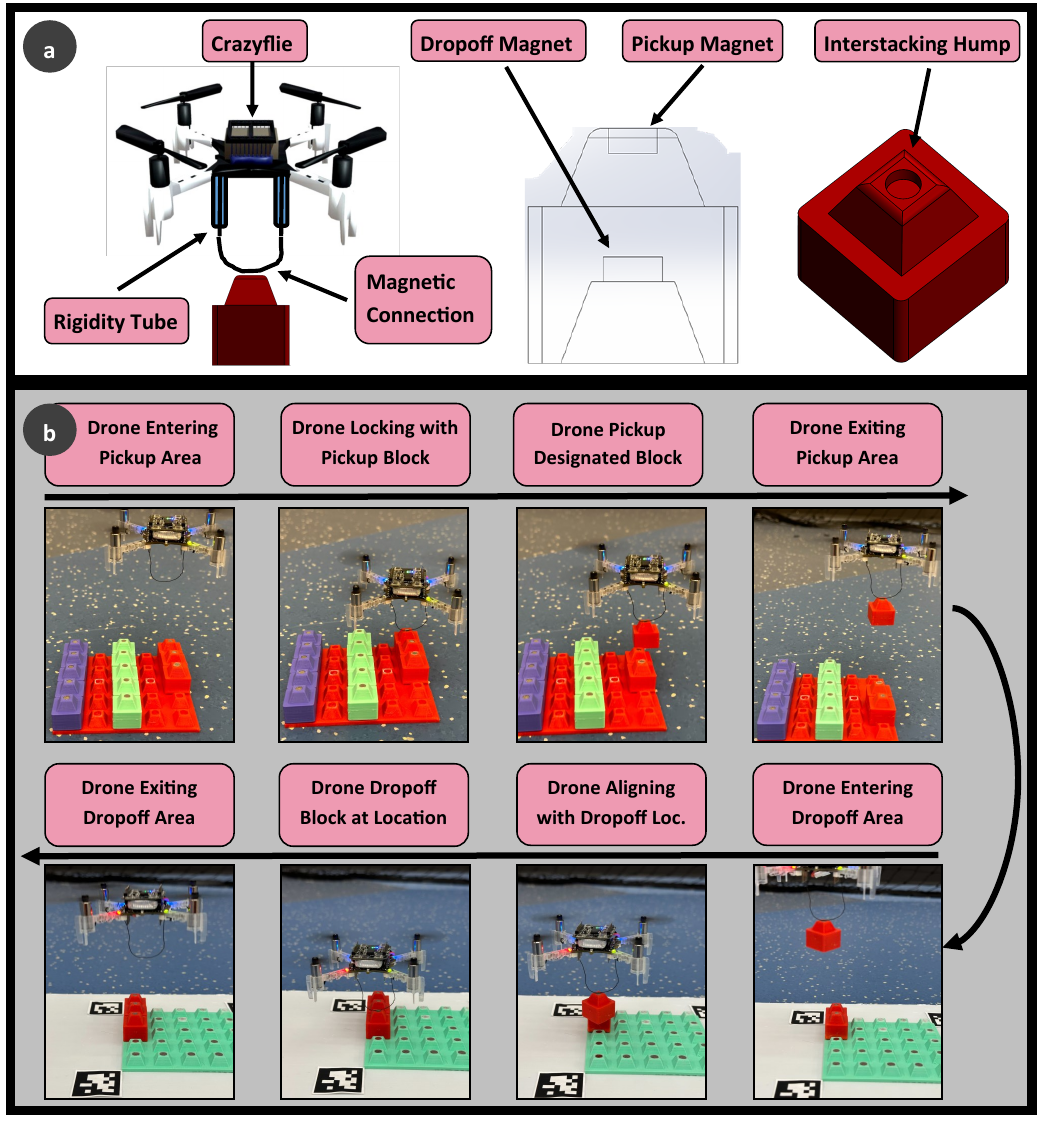}}
\caption{(a) Model of Crazyflie pickup apparatus. A rigid tube keeps the z-length of ferrous wire constant over numerous pickup/dropoff attempts. The wire has the ability to move toward magnetic attraction from the `pickup magnet' on the block. (b) Timelapse of the pickup and dropoff procedure. The stronger dropoff magnet allows the drone to detach from the building block once it is placed.}
\label{fig:B_Mechanical_Module}
\end{figure}

The primary objective of the block design was to develop a viable building system that is within the payload capacity constraints of small drones, such as a Crazyflie. Given the limited payload capacity of these drones, the design needed to be lightweight, compact, and capable of integrating well with other blocks to facilitate complex structures \cite{crazyflie21_datasheet}. The goal was to create a system that allows for simple pickup and drop-off mechanisms with minimal parts, making it feasible for small drones to operate efficiently. For larger drones, this concept can be expanded with more robust features, leveraging the additional payload capacity available.

\subsubsection{Block Shape and Connectivity}

  The block design was inspired by the interlocking capabilities of magnets. Magnets provide a straightforward and effective method for blocks to connect, allowing for relative inaccuracy in placement while still achieving a secure fit. This feature is particularly beneficial in aerial manufacturing, where maintaining stable hover positions without excessive movement is challenging. Magnetic interconnection ensures that the blocks can connect to each other with ease, even if the positioning of the drone is not perfectly precise \cite{cho2013novel, 8632705}.

Each block features a magnetized interlocking mechanism that facilitates easy alignment and secure attachment. The blocks are designed with a hump at the top, serving as an auto-localizing feature to guide the alignment when stacking blocks. Figure \ref{fig:B_Mechanical_Module} shows the shape and structure of the block.

\subsubsection{Magnetic Transport}
The block pickup mechanism leverages the magnet-based autolocking system to facilitate block handling. Each side of the drone incorporates a rotatable wire acting as a hinge, allowing movement with one degree of freedom. The wire hangs free below the drone and rotates towards the magnetic attraction of the block when near. Figure \ref{fig:B_Mechanical_Module} highlights the key components of the mechanism below the drone. Once the block is locked, the drone can carry and position it at the desired location. A stronger magnet, integrated into the drop-off system, attracts the block more forcefully, allowing the drone to detach from the placed block.

\section{Experiments}
\subsection{Testing Environments}

To compare the robustness and overall feasibility of the build environment, as well as the LLM's capabilities, our system was evaluated with two sets of experiments: the first examined the LLM creativity and robustness to errors during manufacturing, and the second examined the systems ability to take in commands from the LLM and convert them into correct actions and provide valid feedback to remedy any errors. Essentially, we design a virtual test scenario that validates the LLM performance and another physical scenario to validate hardware cooperation and the ability to utilize LLM actions.

\subsection{Metrics}
We tested three different LLMs: Claude 3.5 Sonnet, OpenAI 4-o \cite{electronics13132657}, and Gemini Pro 1.5\cite{geminiteam2024gemini15unlockingmultimodal}, comparing their performance in both quantitative and qualitative tests.

The first test was a quantitative assessment of LLM designs on a 10x10 grid. For this test, we created fifteen ``constrained prompts" - prompts in which only one answer can be considered correct. The purpose of these constrained prompts was to restrict the output of the LLMs for precise accuracy measurement. We evaluated their performance using an Intersection over Union (IoU) metric, comparing the generated designs with manually defined correct answers. Each LLM was tested five times per prompt and an average IoU was calculated across these responses. Figure \ref{fig:D_LLM_Performance_Results}a provides examples of six constrained prompts, their corresponding correct design plots, and the average IoU for each model on those specific prompts. Figure \ref{fig:D_LLM_Performance_Results}b provides the average IoU for the 15 prompts, the variance, the inference time in milliseconds, and the cost per 1000 tokens for each model.


The second test was a qualitative comparison conducted on a smaller 5x5 grid. This test focused on open-ended design requests for simple geometric shapes, such as ``star", ``trapezoid", and ``right triangle". We assessed the feasibility and recognizability of the generated designs for each request. Feasibility was determined by whether the LLM followed the given guidelines, including staying within the 5×5 grid, using only integer values for coordinates, and responding in the correct JSON format, all of which were specified in the prompt. Recognizability was judged based on whether a person could correctly identify the intended shape without prior knowledge of the design request. Human evaluators graded each design based on these criteria, using a three-point scale: 1 indicated the design was both feasible and recognizable, 2 indicated it met only one of these criteria, and 3 indicated it met neither.  Figure \ref{fig:D_LLM_Performance_Results}C shows a bar chart showing the number of times each LLM scored 1, 2, or 3 on 30 different designs that we evaluated. 


\subsection{LLM Results}
\begin{figure}[h!]
\centerline{\includegraphics[width=\textwidth]{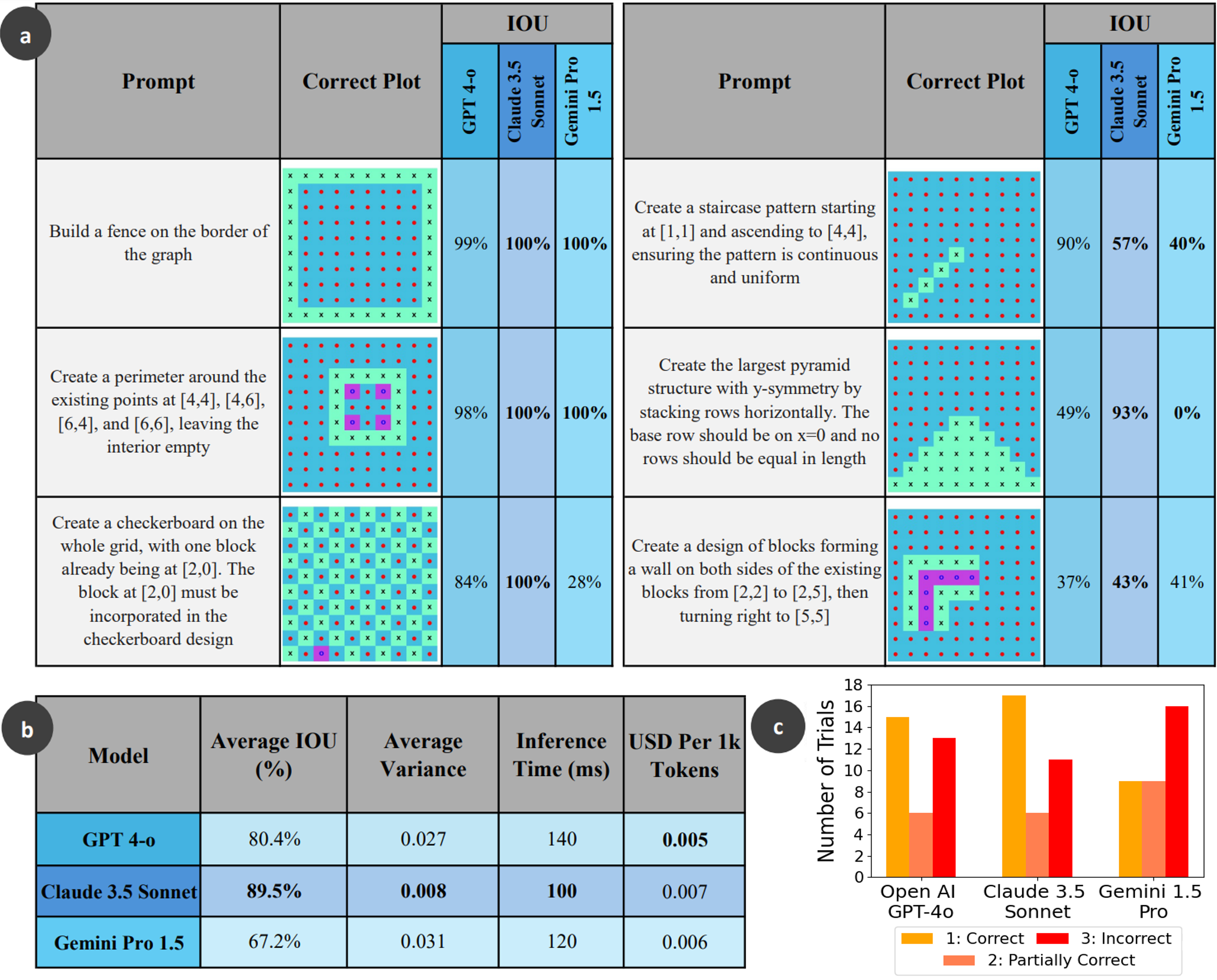}}
\caption{Performance of LLMs in the quantitative test (a and b) and qualitative test (c).}
\label{fig:D_LLM_Performance_Results}
\end{figure}

The quantitative results of our test, shown in Figures \ref{fig:D_LLM_Performance_Results}a and \ref{fig:D_LLM_Performance_Results}b, show that Claude 3.5 Sonnet achieved the highest average IoU at 89.5\% in the fifteen constrained prompt tests. Furthermore, Claude 3.5 Sonnet exhibited the lowest average variance at 0.008, indicating greater consistency in its responses, regardless of whether the answers were correct or incorrect. GPT 4-o also performed well in spatial reasoning tasks, with an average IoU of 80.4\%, but showed a higher average variance of 0.027. In contrast, Gemini Pro 1.5 had the lowest average IoU at 67.2\%, coupled with the highest average variance at 0.031. Overall, our results show that the best-performing model, Claude 3.5 Sonnet, excelled due to its ability to both generate correct designs and maintain consistency across trials of the same design request.

The qualitative results, shown in Figure \ref{fig:D_LLM_Performance_Results}c, reveal that the designs generated by Claude 3.5 Sonnet were rated slightly higher than those of GPT 4-o, and both models outperformed Gemini Pro 1.5 by a significant margin. These findings suggest that both Claude 3.5 Sonnet and GPT 4-o demonstrate the capability to handle open-ended design tasks. 

Overall, we find that Claude 3.5 Sonnet demonstrates the greatest ability for spatial reasoning within the framework of our task. Additionally, it has the fastest inference time of the three models, further enhancing its suitability for time-sensitive tasks, though at a slightly higher cost.

\subsection{Crazyflie Ecosystem Performance}
\begin{figure}[h!t]
    \centering
    \centerline{\includegraphics[width=\textwidth]{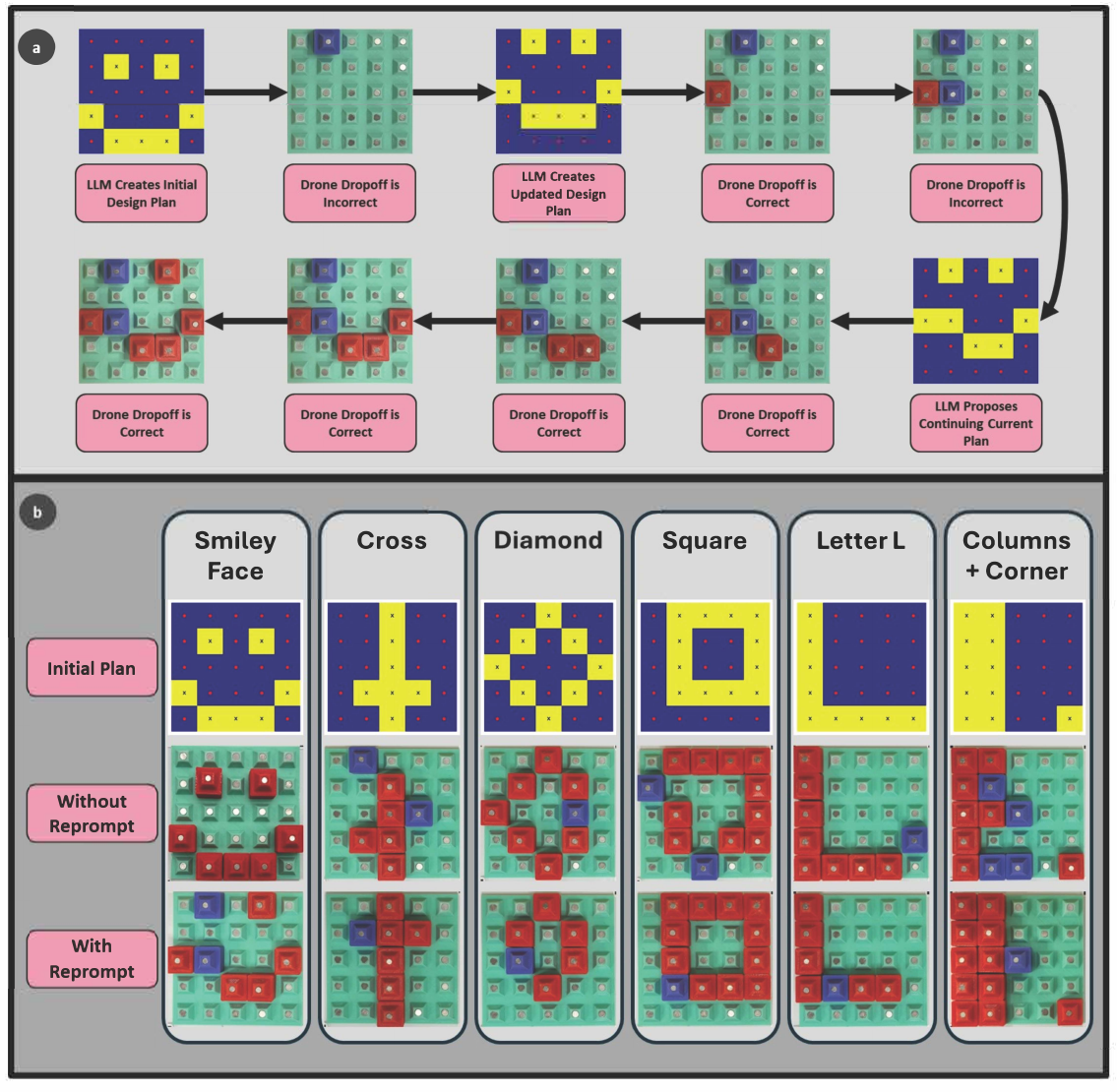}}
    \caption{Crazyflie Ecosystem Results: a) Outlines a step-by-step design process with reprompting enabled for the design of a Smiley Face. b) Provides an overview of 6 designs both with and without reprompting.}
    \label{fig:E_Crazyyflie_Ecosystem_Results}
\end{figure}

We tested the LLM-Drone pipeline with the Crazyflie ecosystem on a 5x5 grid build space. We executed the pipeline for several design requests, including a smiley face, cross, diamond, square, the letter L, and ``two columns on the left and bottom right corner only." In order to evaluate the effectiveness of reprompting the LLM on placement errors by the drone, we execute runs for each of these designs both with and without reprompting enabled.
Figure \ref{fig:E_Crazyyflie_Ecosystem_Results}a provides a step-by-step walkthrough of the pipeline executing the design of a smiley face with reprompting enabled. Figure \ref{fig:E_Crazyyflie_Ecosystem_Results}b provides a comparison of the initial LLM design plan (top), an execution without reprompting (middle), and an execution with reprompting (bottom). Note that in all images we manually replaced incorrectly placed blocks with purple blocks after the experiment concluded to aid in visualization of where errors occurred.
The results of reprompting versus no reprompting show that many designs can be recovered by the LLM after the drone incorrectly places a block. For example, when tasked with designing a cross, the LLM identifies that it can simply rotate the coordinates to accommodate the misplaced block. For the diamond and letter L designs, the LLM provides new layouts that incorporate misplaced blocks. Finally, for the square design, the LLM identifies that the misplaced block was simply an out-of-order placement, thus keeping the same overall design while telling the drone to return to the spot it previously missed. In general, these tests show that many errors can be overcome by providing the LLM with information about the current scene and allowing it to update the design to accommodate various errors.

\section{Conclusion}
The integration of Large Language Models (LLMs) with aerial additive manufacturing represents a transformative step forward in the field of construction and logistics, offering a dynamic solution to the challenges posed by inaccessible terrains and complex construction tasks. Through our research, we demonstrate the capabilities of LLM-driven drones to achieve high levels of precision and adaptability in real world settings. Our findings indicate that the LLM Planner can manage up to 90\% build accuracy, enhancing both the planning and execution phases of aerial manufacturing tasks. We also conduct a performance comparison across various LLM platforms, revealing that Claude 3.5 Sonnet outperformed others in spatial reasoning and subjective correctness of object construction.

Our research underscores the great potential of merging cognitive computing with physical manufacturing processes. The LLM-Drone system paves the way for innovative applications in various industries, including remote construction, warehouse automation, and emergency infrastructure deployment. Future work can explore using LLM reasoning to construct multilayer structures, allowing complex 3D objects to be built. On a mechanical level, the system is designed to be adaptable, with the potential for enhancements such as larger, more capable drones and the incorporation of magnets that can be turned on and off. These changes would further optimize the performance and flexibility of the design, allowing for more precise control and a wider range of applications \cite{cho2013novel, 8632705}.





\bibliography{references}

\newpage
\appendix
\section{Appendix A: Example LLM Prompt and Response}
\label{appendix:LLMPrompt}
\begin{figure}[h!t]
    \centering
    \centerline{\includegraphics[width=\textwidth]{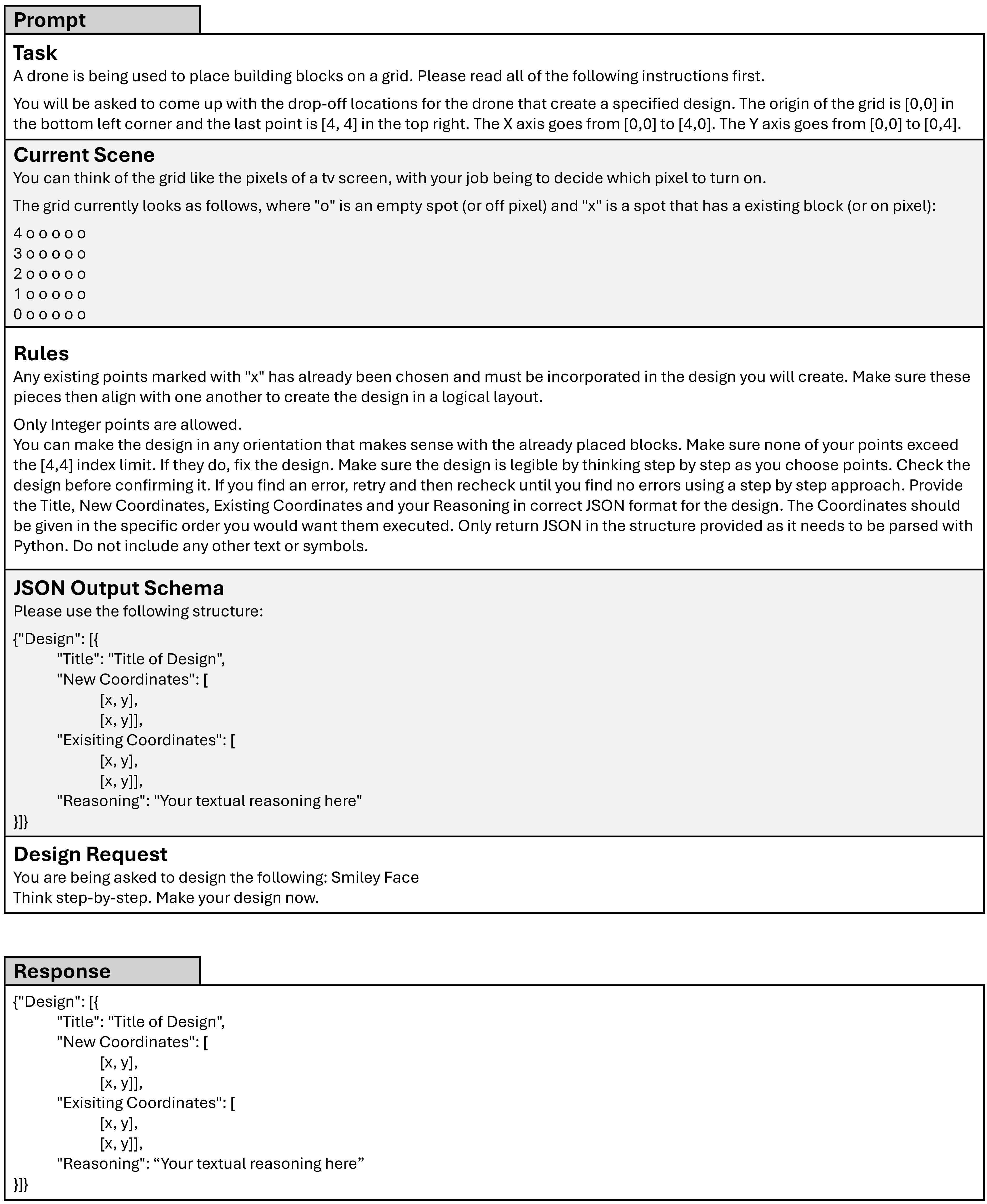}}
\end{figure}

\label{sec:appendix}

\end{document}